\documentclass[letterpaper]{article} % DO NOT CHANGE THIS
\usepackage{aaai25}  % DO NOT CHANGE THIS
\usepackage{times}  % DO NOT CHANGE THIS
\usepackage{helvet}  % DO NOT CHANGE THIS
\usepackage{courier}  % DO NOT CHANGE THIS
\usepackage[hyphens]{url}  % DO NOT CHANGE THIS
\usepackage{graphicx} % DO NOT CHANGE THIS
\usepackage{natbib}  % DO NOT CHANGE THIS AND DO NOT ADD ANY OPTIONS TO IT
\usepackage{caption} % DO NOT CHANGE THIS AND DO NOT ADD ANY OPTIONS TO IT
\usepackage{algorithm}
\usepackage{algorithmic}

\usepackage{amsmath}
\usepackage{amssymb}
\usepackage{algorithm}
\usepackage{algorithmic}
\usepackage{svg}
\usepackage{multirow}
\usepackage{diagbox}
\usepackage{adjustbox}
\usepackage{tabularx}
\usepackage{rotating}
\usepackage{booktabs}
\usepackage{xcolor}
\usepackage{colortbl}
\usepackage{xr}
\usepackage{url}

\usepackage{newfloat}
\usepackage{listings}
\DeclareCaptionStyle{ruled}{labelfont=normalfont,labelsep=colon,strut=off} % DO NOT CHANGE THIS
\floatstyle{ruled}
\newfloat{listing}{tb}{lst}{}
\floatname{listing}{Listing}
\title{Enhance Vision-Language Alignment with Noise}
\author{
    Sida Huang\textsuperscript{\rm 1,2},
    Hongyuan Zhang\textsuperscript{\rm 2, 3}\thanks{Corresponding author.},
    Xuelong Li\textsuperscript{\rm 2}\footnotemark[1]
}
\affiliations{
    \textsuperscript{\rm 1}Northwestern Polytechnical University\\
    \textsuperscript{\rm 2}Institute of Artificial Intelligence (TeleAI), China Telecom\\
    \textsuperscript{\rm 3}The University of Hong Kong\\
    \{sidahuang2001, hyzhang98\}@gmail.com, xuelong\_li@ieee.org
}

\usepackage{bibentry}
\begin{document}

\maketitle

\begin{abstract}
With the advancement of pre-trained vision-language (VL) models, enhancing the alignment between visual and linguistic modalities in downstream tasks has emerged as a critical challenge.
Different from existing fine-tuning methods that add extra modules to these two modalities, we investigate whether the frozen model can be fine-tuned by customized noise. 
Our approach is motivated by the scientific study of beneficial noise, namely Positive-incentive Noise (Pi-noise or $\pi$-noise) , which quantitatively analyzes the impact of noise. 
It therefore implies a new scheme to learn beneficial noise distribution that can be employed to fine-tune VL models.
Focusing on few-shot classification tasks based on CLIP, we reformulate the inference process of CLIP and apply variational inference, demonstrating how to generate $\pi$-noise towards visual and linguistic modalities.
Then, we propose \textbf{P}ositive-\textbf{i}ncentive \textbf{N}oise \textbf{I}njector (PiNI), which can fine-tune CLIP via injecting noise into both visual and text encoders.
Since the proposed method can learn the distribution of beneficial noise, we can obtain more diverse embeddings of vision and language to better align these two modalities for specific downstream tasks within limited computational resources.
We evaluate different noise incorporation approaches and network architectures of PiNI. The evaluation across 11 datasets demonstrates its effectiveness.
Our code is available at: \url{https://github.com/hyzhang98/PiNI}.
\end{abstract}

\section{Introduction}
Vision and language are two crucial modalities in the real world. 
To build association and enable collaboration between these two modalities, vision-language (VL) models have emerged in recent years.
Based on the classical dual-stream VL model CLIP \cite{CLIP}, LLaVA \cite{llava, llava1_5} has recently achieved remarkable performance.
In these models, the embeddings of these two modalities are aligned during the pre-training phase.
When applying CLIP models to downstream unseen data, \textit{inherent dataset bias can lead to misalignment between visual and linguistic representations}. 
To realign these two modalities, fine-tuning all parameters requires extensive computational resources.
Even with sufficient computational resources, full fine-tuning on limited downstream data also carries a risk of overfitting and forgetting the knowledge acquired during pre-training stage.
Consequently, it is crucial to effectively and efficiently fine-tune VL models to achieve better alignment.

To address this challenge, existing methods such as prompt tuning \cite{Prompt_Tuning} and adapter tuning \cite{Adapter} are proposed.
Prompt tuning sets the word embeddings of prompts as learnable parameters and adapter tuning inserts a learnable shallow neural network to models.
Different from these fine-tuning methods that add extra modules or concatenate extra tokens, we focus on an interesting question:
\textbf{\textit{Can a frozen model be tuned with customized noise?}}

It is motivated by Positive-incentive Noise (Pi-noise or $\pi$-noise) \cite{pi-noise, VPN}, which quantitatively analyzes the impact of noise and investigates how to generate beneficial noise for a specific task. 
Different from traditional concepts that treat the randomness of noise as a harmful factor, the noise $\mathcal{E}$ can decrease the uncertainty of predictions and simplify the task if it satisfies
\begin{equation}
    I(\mathcal{T}, \mathcal{E}) > 0 \Leftrightarrow H(\mathcal{T}) > H(\mathcal{T}|\mathcal{E}),
    \label{eqn_definition}
\end{equation}
where $\mathcal{T}$ is the definition of a specific task from a probabilistic perspective, $I(\cdot,\cdot)$ denotes mutual information and $H$ represents entropy.
It should be clarified that our method is distinct from noise addition in data augmentation, as discussed in Section \ref{subsec_augmentation}.  

\begin{figure}[t]
    \centering
    \includegraphics[width=\linewidth]{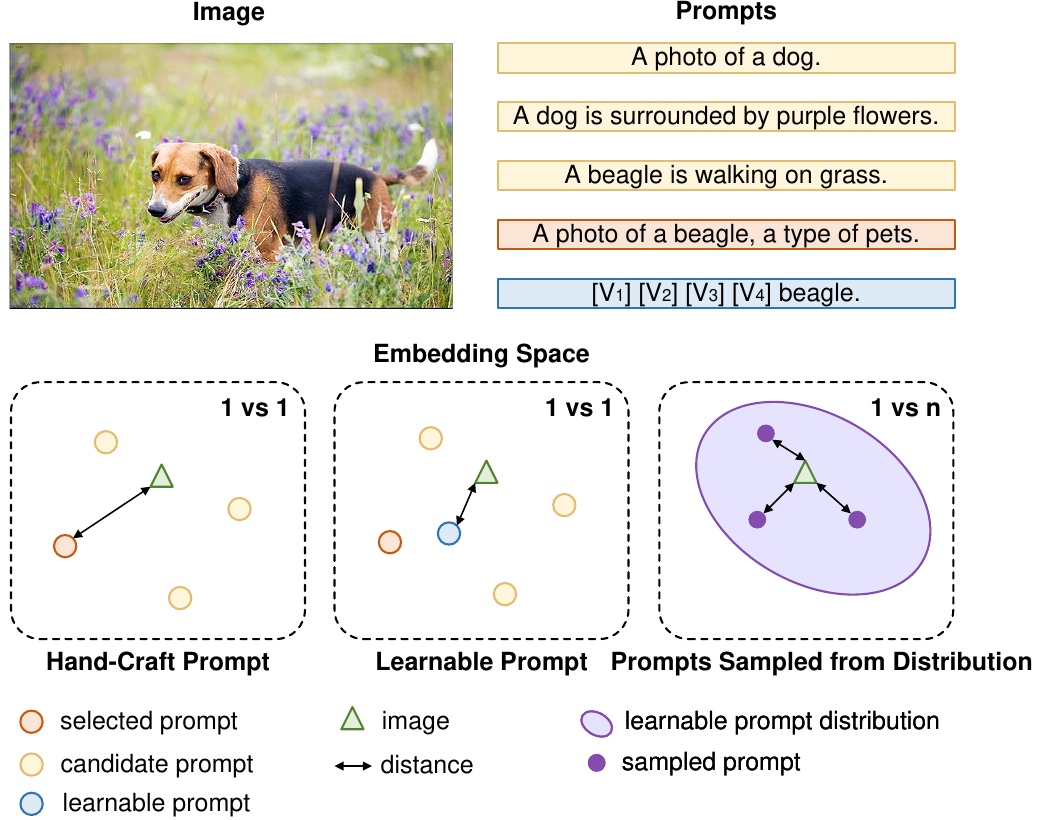}
    \caption{Different strategies for constructing prompts. 
    A single image can be described by multiple prompts. 
    The first and second strategies each correspond to one prompt per image.
    After fine-tuning, the learnable prompt becomes closer to the image in the embedding space.
    After learning the noise distribution in prompts, we can easily sample prompts with richer and more precise semantics as needed.
    }
    \label{fig_strategies_prompts}
\end{figure}

The advantages of noise to enhance alignment can be discussed from both vision and language perspectives.
\textbf{From the perspective of vision}, the biases in different visual datasets \cite{dataset_bias} can lead to performance degeneration in VL models.
% It can be mitigated through learning the noise distribution of this bias.
We hypothesize that the bias associated with each image follows a certain distribution, which can be mitigated by adding noise also from a specific distribution.
As shown in Figure \ref{fig_visual}, the images with added noise exhibit more universal visual features.
Additionally, the limited number of images makes it difficult to adequately sample from the continuous space of image embeddings. 
This dilemma can be alleviated by adding randomly sampled beneficial noise into the visual level, for which more diverse visual embeddings can be obtained. 
\textbf{From the perspective of language}, the templates of hand-crafted prompts and learnable prompts may be not effective enough, which limit the diversity of linguistic modality. 
Figure \ref{fig_strategies_prompts} shows two strategies and our proposed scheme for prompt construction. In the original CLIP \cite{CLIP}, the hand-crafted prompt is used, e.g., a typical form ``a photo of a [class]''. Some prompt tuning works \cite{CoOp, Co_CoOp, LAMM} replace the hand-crafted prompt with trainable word embeddings to mine more precise description of images. 
These prompts are class-instanced. In other words, a class label corresponds to only one prompt.
Nevertheless, a class should be related to various prompts (with different qualities). 
By sampling different beneficial noise from a customized distribution and adding it to hand-crafted prompts, new prompts can be easily obtained. 
It is thus easy to find that the prompts should also obey some probability distribution, ensuring the semantic richness of prompts.

Towards prompt-based image classification task, we propose PiNI based on $\pi$-noise theory which can enhance the alignment of VL models. 
Overall, the contributions can be summarized as follows:

\begin{itemize}
  \item We propose \textbf{P}ositive-\textbf{i}ncentive \textbf{N}oise \textbf{I}njector (PiNI), a fine-tuning method for CLIP through injecting customized beneficial noise into it. To the best of our knowledge, this is the first noise-based approach to fine-tune CLIP.
  \item We reformulate the inference process of CLIP, by treating the prompt as a variable to analyze the effect of both images and prompts on the probability of category labels.
  It therefore offers a paradigm to refine CLIP from a probabilistic aspect. 
  \item We apply variational inference and Monte Carlo method to converting the complex loss with noise distribution to a tractable one, which guides us in generating and injecting noise.
  Extensive experiments validate our idea.
\end{itemize}

\section{Related Works}
\subsection{Fine-Tuning Methods for VL Models}
It is crucial to fine-tune VL models for downstream tasks with limited computational resources.
To address this challenge, some Parameter-Efficient Fine-Tuning (PEFT) methods have been proposed, primarily including parameter tuning, adapter tuning \cite{Adapter} and prompt tuning \cite{Prompt_Tuning}.
The parameters of VL model are directly tuned in parameter tuning. 
LoRA \cite{LoRA} and Bitfit \cite{Bitfit} focus on tuning the weights and biases of models, respectively.
Adapter tuning inserts a trainable module named Adapter into a frozen model. CLIP-Adapter \cite{CLIP-Adapter} and VL-Adapter \cite{VL-Adapter} utilize Adapter in fine-tuning VL models.
Prompt tuning was first proposed in NLP, which adds a learnable prompt template to  pre-trained language models. Since CLIP model includes a text encoder, numerous works have attempted to introduce prompt tuning to CLIP \cite{CoOp, Co_CoOp, LAMM, CALIP}. Beyond text encoders, VPT \cite{VPT} and MaPLe \cite{MaPLe} extend prompt tuning to visual encoders. 
Different from the above works, we focus on learning noise distribution to sample more visual and text embeddings without altering the architecture of VL models.

\subsection{Difference with Data Augmentation}
\label{subsec_augmentation}
Adding noise during training as a data augmentation strategy \cite{learning_augmentation} is common in computer vision tasks, such as adversarial training \cite{adversarial_segmentation, adversarial_style_aug, augment_uncertainty}.
The data augmentation only participates in the training phase and is excluded during the inference phase. 
However, our proposed method injects noise for both the training and inference phases.
For adversarial augmentation, it aims at adding perturbations to the input of networks, thereby enhancing robustness. 
Conversely, the noise in PiNI is employed to simplify tasks instead of creating difficulty for base models.

\section{Method}
When applying VL models to unseen downstream data, dataset bias \cite{dataset_bias} can lead to misalignment in representations of vision and language.
To mitigate this issue, we focus on achieving better vision-language alignment in the prompt-based image recognition task $\mathcal{T}$. 
Inspired by $\pi$-noise \cite{pi-noise, PiNDA} , we propose a new scheme for learning beneficial noise distribution, namely \textbf{P}ositive-\textbf{i}ncentive \textbf{N}oise \textbf{I}njector (PiNI).

In this section, we first reformulate the inference process of CLIP in Section \ref{subsec_clip}. 
Section \ref{subsec_pi_noise} introduces how to define the task entropy on the specific dataset $\mathcal{X}$, which is vital to calculate the optimization objective $I(\mathcal{T},\mathcal{E})$ when generating $\pi$-noise.
To optimize this objective, the variational approximation is applied to getting its upper bound in Section \ref{subsec_vari_appro}.
In the final approximate loss, we adopt a learnable function to approximate noise and determine the noise incorporation approaches, which are discussed in Section \ref{subsec_noise_fine_tune}.

\subsection{Reformulation of CLIP Inference} \label{subsec_clip}

Different from classical image classification models that only require images as input, CLIP additionally needs a prompt set to obtain the position of each category in the embedding space.
Obviously, the inference process of CLIP relies on the prompts. 
However, in traditional models,  the label probability is usually modeled as $p(y|x)$, which ignores the role of prompts.

It is rational to introduce a separate variable for prompts.
Given an image $x$ and a set of prompts $\rho_i \in \mathcal{P}, i \in {1,2,...,N}$, the probability of class  $y \in {1,2,...,N}$ can be reformulated as
\begin{equation} \label{eqn_clip}
  p(y|x, \mathcal{P}) = \frac{\exp(\textrm{sim}(V(x), T(\rho_y))/\tau)} {\sum_{i=1}^{N} \exp(\textrm{sim}(V(x), T(\rho_i))/\tau)},
\end{equation}
where $V$ and $T$ represent the visual and text encoders, $\textrm{sim}(\cdot, \cdot)$ is the similarity between two vectors, and $\tau$ is a temperature parameter. 
Note that $\mathcal{P}$ is a variable, and there exists a one-to-one correspondence between prompt  $\rho_i$ and label $y$. 
In CLIP model, $\textrm{sim}(\cdot, \cdot)$ is the cosine similarity. 

\subsection{$\boldsymbol{\pi}$-noise Regarding Prompts}
\label{subsec_pi_noise}
Primarily, it is crucial to evaluate the complexity of a classification task on an arbitrary dataset $\mathcal{X}$ and prompt set $\mathcal{P}$. Inspired by \cite{pi-noise}, task entropy can be used to formulate the complexity as
\begin{equation}
    \begin{aligned}
        H(\mathcal{T}) 
        & = H(y|x, \mathcal{P})\\
        & = \mathbb{E}_{x \sim \mathcal{D}_\mathcal{X}} \mathbb{E}_{y \sim p(y|x, \mathcal{P})}[-\log p(y|x,\mathcal{P})] \\
        & = \mathbb{E}_{p(x,y| \mathcal{P})}[-\log p(y|x,\mathcal{P})], \\
    \end{aligned} \label{eqn_HT}
\end{equation} 
where $\mathcal{D}_\mathcal{X}$ is the data distribution over $\mathcal{X}$.  
This equation measures complexity by computing the uncertainty of labels. 
For a given image, it may resemble multiple categories, which increases the uncertainty of the label $y$. The noise $\mathcal{E}$ can decrease this uncertainty if it satisfies Eq. (\ref{eqn_definition}).
The definition of $I(\mathcal{T}, \mathcal{E})$ is
\begin{equation}
    I(\mathcal{T},\mathcal{E}) =  H(\mathcal{T}) - H(\mathcal{T} | \mathcal{E}).
\end{equation}
Similar to $H(\mathcal{T})$, $H(\mathcal{T} | \mathcal{E})$ is defined as 
\begin{equation}
    \begin{aligned}
        H(\mathcal{T} | \mathcal{E}) 
        & = \mathbb{E}_{p(x,y,\varepsilon | \mathcal{P})}[- \log p(y|x, \varepsilon, \mathcal{P})]. \\
        % & = \mathbb{E}_{x, y|\mathcal{P}} \mathbb{E}_{\varepsilon|x,y,\mathcal{P}} [- \log p(y|x, \varepsilon, \mathcal{P})]
    \end{aligned} \label{eqn_condition_entropy}
\end{equation}
As $H(\mathcal{T})$ is a constant term for fixed CLIP model, maximizing $ I(\mathcal{T},\mathcal{E})$ is equivalent to minimizing $H(\mathcal{T} | \mathcal{E})$.

\subsection{Variational Inference for the Intractable Problem
}
\label{subsec_vari_appro}
\begin{figure}[t]
    \centering
    \includegraphics[width=0.9\linewidth]{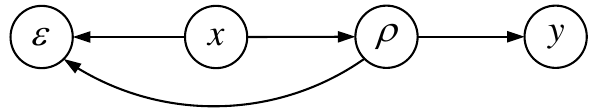}
    \caption{The probabilistic graphical model of PiNI.   
    }
    \label{fig_pro_graph}
\end{figure}
Since $p(y|x, \varepsilon, \mathcal{P})$ is difficult to calculate precisely due to the integral, we apply the variational inference technique \cite{variational_inference} to Eq. (\ref{eqn_condition_entropy}) and obtain a variational upper bound of $H(\mathcal{T}|\mathcal{E})$ as 
\begin{equation}
    \begin{aligned}
        \mathcal{L} =
        & \mathbb{E}_{p(x, y, \varepsilon|\mathcal{P})} [- \log q(y|x, \varepsilon, \mathcal{P})]
        \geq H(\mathcal{T} | \mathcal{E}) , \\ 
    \end{aligned}  \label{upper_bound}
\end{equation} 
where $q(y|x, \varepsilon, \mathcal{P})$ is a tractable approximation of $p(y|x, \varepsilon, \mathcal{P})$.
The above inequality is derived from the non-negative property of KL-Divergence as follows, 
\begin{equation}
    KL(p||q) \geq 0 \Leftrightarrow \mathbb{E}_{p(x)}[\log p(x)] \geq \mathbb{E}_{p(x)}[\log q(x)].
\end{equation}
The current problem is how to effectively compute this variational bound.
Monte Carlo estimation can be applied to data distribution $\mathcal{D}_\mathcal{X}$ to obtain image-label pairs $(x_i,y_i)$. So the variatonal bound is further approximated  as
\begin{equation}
    \mathcal{L} \approx \frac{1}{n} \sum_{i=1}^{n} \mathbb{E}_{p(\varepsilon|x_i, y_i,\mathcal{P})} [- \log q(y_i|x_i, \varepsilon, \mathcal{P})].
\end{equation}
For further calculation, $p(\varepsilon|x_i, y_i,\mathcal{P})$ needs to be simplified as there are three given variables. 
In Figure \ref{fig_pro_graph}, the probabilistic graphical model illustrates the dependencies among variables. 
The noise and the label are conditionally independent given the prompt: $\varepsilon \perp y \, | \, \rho$. 
Accordingly, we can obtain the following result from the conditional independence, 
\begin{equation} \label{eqn_conditional_independence}
    p(\varepsilon|x, y, \rho) = p(\varepsilon|x, \rho).
\end{equation}
The above derivations indicate that the generated noise should be controlled by the set of prompts so that it can exclude interference from information irrelevant to prompts. 
Therefore,  the variational bound is further transformed to 
\begin{equation}
    \mathcal{L} \approx \frac{1}{n} \sum_{i=1}^{n} \mathbb{E}_{p(\varepsilon|x_i, \mathcal{P})} [- \log q(y_i|x_i, \varepsilon, \mathcal{P})]. \label{eqn_loss_conti}
\end{equation}
We assume that $p(\varepsilon|x, \mathcal{P})$ follows a classical Gaussian distribution, whose distribution parameters $\mu$ and $\Sigma$ is approximated using a learnable function $f_\theta(x_i, \mathcal{P})$:
\begin{equation}
(\mu, \Sigma) = f_\theta(x_i, \mathcal{P}),
\label{eqn_mu_Sigma}
\end{equation}
where $\theta$ represents the parameters of the function.
However, calculating the integral of the continuous variable $\varepsilon$ is difficult. 
The Monte Carlo method is applied again to address this problem. 
To ensure the back propagation of gradients in $f_\theta(x_i, \mathcal{P})$, the reparameterization trick \cite{VAE} introduces an auxiliary variable $\epsilon \sim N(0,I)$ satisfying $p(\varepsilon|x_i, \mathcal{P})d\varepsilon = p(\epsilon)d\epsilon$.
Subsequently, a learnable function $G_\theta$ is established to generate noise
\begin{equation}
\varepsilon = G_\theta(\epsilon, x_i, \mathcal{P}) = \Sigma_\theta(x_i, \mathcal{P}) \cdot \epsilon + \mu_\theta(x_i, \mathcal{P}).
\label{eqn_G_theta}
\end{equation}
We randomly sample $\epsilon$ $m$ times for each $x_i$ to get the final loss function for training
\begin{equation} 
    \begin{aligned}
        \mathcal{L} 
        & \approx \frac{1}{n} \sum_{i=1}^{n} \mathbb{E}_{p(\epsilon)}[- \log q(y_i|x_i, G_\theta(\epsilon, x_i, \mathcal{P}), \mathcal{P})] \\
        & \approx \frac{1}{n \cdot m} \sum_{i=1}^{n} \sum_{j=1}^{m} [- \log q(y_i|x_i, G_\theta(\epsilon_{ij}, x_i,  \mathcal{P}), \mathcal{P})].  \label{eqn_loss}
    \end{aligned}
\end{equation} 

\begin{figure*}[!t]
    \centering
    \includegraphics[width=\linewidth]{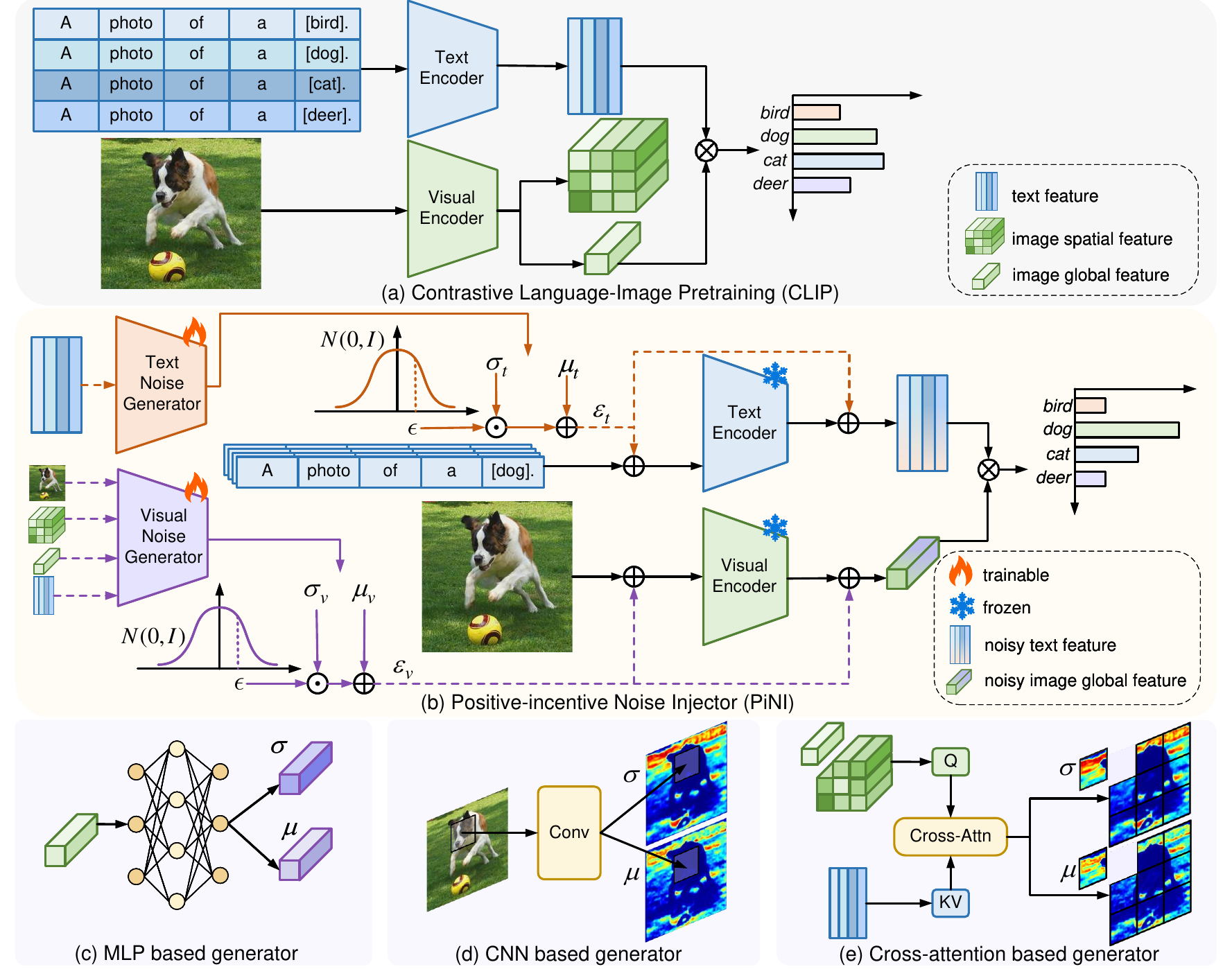}
    \caption{\textbf{(a)} The inference process of CLIP, whose output features are utilized in PINI. 
    \textbf{(b)} The framework of PiNI, which includes two procedures for adding noise.
    There are multiple options for the input of noise generators and locations of noise injection, whose potential transmission paths are represented with dashed lines. 
    In the figure, $\odot$ denotes the Hadamard product, $\oplus$ denotes matrix or vector addition, and $\otimes$ denotes matrix multiplication.
    \textbf{(c)(d)(e)} are three architectures of noise generators for learning distribution parameters. 
    }
    \label{fig_framework}
\end{figure*}

\subsection{Fine-Tuning with $\boldsymbol{\pi}$-noise} \label{subsec_noise_fine_tune}
In the derived final loss Eq. (\ref{eqn_loss}), there are two important components: $q(y|x, \varepsilon, \mathcal{P})$ and $\varepsilon = G_\theta(\epsilon, x, \mathcal{P})$. 
The former predicts the labels with the aid of noise. 
The specific approaches for incorporating noise will be discussed below. 
The latter $G_\theta(\epsilon, x, \mathcal{P})$ defined in Eq. (\ref{eqn_G_theta}) consists of two phases: (1) Sampling: $\epsilon \sim N(0,I)$ and (2) Distribution parameter estimation: $(\mu, \Sigma) = f_\theta(x, \mathcal{P})$. 
$f_\theta(x, \mathcal{P})$ can be implemented using various neural network architectures, which will be introduced in the following discussion. The detailed framework of PiNI is illustrated in Figure \ref{fig_framework}.

\begin{figure*}[!t]
    \centering
    \includegraphics[width=\linewidth]{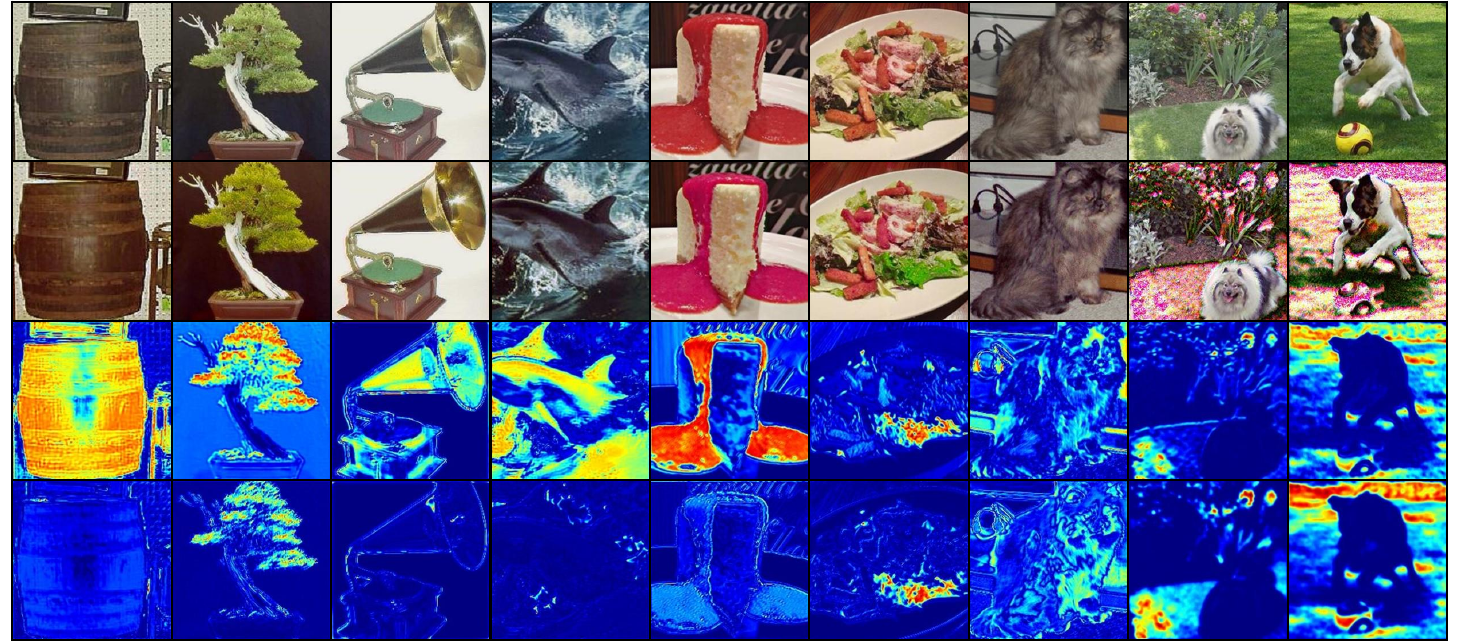}
    \caption{Visualization of generated noise injected into raw images.
    The first row shows the raw images. The second row displays the noise-injected images. The third and fourth rows present the heatmaps of the mean $\mu$ and variance $\sigma$ for each pixel, respectively.
    In the first column, the noise deepens the color of an old barrel, making it look new again and thereby reducing the bias between datasets.
    The vegetation and ball in the last column are disturbed by noise, simplifying the task of recognizing.
    }
    \label{fig_visual}
\end{figure*}

\subsubsection{Noise Incorporation Approach}
% 1. input no change
% 2. 
% 3. noise in prompt and image 
% 4. feature and word embedding
When approximating $q(y|x, \varepsilon, \mathcal{P})$ to predict labels, 
% our motivation is to fully utilize the inference process of CLIP, that is $p(y|x, \mathcal{P})$ in Eq. (\ref{eqn_clip}).
we avoid altering the original inference process of CLIP as formulated in Eq. (\ref{eqn_clip}). 
It will remain the architecture of CLIP.  
A simple approach is to inject noise $\varepsilon$ into the two encoders.   
To simplify the discussion, we employ the classical factorization trick to \textit{decompose the noise into two components} $\varepsilon = \{ \varepsilon_v, \varepsilon_t \}$, where $\varepsilon_v$ and $\varepsilon_t$ are injected into the visual and text encoders, respectively. 
Note that similar to the classical Naive Bayes model \cite{PRML}, a hypothesis that $\varepsilon_v$ and $\varepsilon_t$ are conditional independent is established. 
According to this hypothesis,  $\varepsilon_v$ and $\varepsilon_t$ can be generated and utilized separately.

As shown in Figure \ref{fig_framework}(b), the visual noise $\varepsilon_v$ can be injected into different locations of CLIP, including raw image or visual feature. 
For the text noise $\varepsilon_t$, it can be injected into word embeddings of prompts or text features.
\cite{DGNN} proposes to train multiple modules simultaneously in GNN.
We will compare the effects of different injection locations in Section \ref{subsec_explore}.
Note that the noise has the same shape as the input at the injection location.
Additionally, additive noise and multiplicative noise correspond to different methods of fusing input and noise.
Unless specifically stated, we use additive noise by default. 

\subsubsection{Architecture of Noise Generator}
%pending
Now, we discuss different implementations of $f_\theta(x, \mathcal{P})$ in Eq. (\ref{eqn_mu_Sigma}) to estimate the distribution parameters.
To maintain the computational efficiency, we simply assume that $\varepsilon$ follows uncorrelated multivariate Gaussian distribution, i.e., $\Sigma$ is a diagonal matrix.  
The learnable variance parameters are represented as $\sigma = \mathrm{Diag}(\Sigma)$.
Based on the above hypothesis, we separately generate $\varepsilon_v$ and $\varepsilon_t$ using $f_v(x, \mathcal{P})$ and $f_t(x, \mathcal{P})$. 
Subsequently, Eq. (\ref{eqn_mu_Sigma}) and Eq. (\ref{eqn_G_theta}) are formulated as
\begin{equation} \label{eqn_standard_gauss}
    \begin{aligned}
        & \varepsilon_v = \sigma_v \odot \epsilon_v + \mu_v, 
        &  \mu_v, \sigma_v  = f_v(x, \mathcal{P}); \\
        & \varepsilon_t = \sigma_t \odot \epsilon_t + \mu_t,
        &  \mu_t,\sigma_t = f_t(x, \mathcal{P}).
    \end{aligned}
\end{equation}

\paragraph{\bf Details of Visual Noise Generator $\boldsymbol{f_v}$}
As shown in Figure \ref{fig_framework}(b),  the input to the visual noise generator can be not only raw image $x$ and prompts $\mathcal{P}$ but also their features encoded by CLIP, including text feature, image spatial feature, and image global feature.
In Figures \ref{fig_framework}(c)-(e), we introduce three architectures for $f_v$ based on MLP, CNN and Cross Attention \cite{Transformer}, which can estimate $\mu$ and $\sigma$ at either the raw image level or the feature level. 

\paragraph{\bf Details of Text Noise Generator $\boldsymbol{f_t}$}
When estimating $\mu_t$ and $\sigma_t$ from $f_t(x, \mathcal{P})$, each image $x$ corresponds to a noise distribution over prompts.  
Thus, the prompt set with added noise should be image-instanced.
Encoding these image-instanced prompts will cause a considerable computational burden. 
Conversely, for the classical CLIP based classification in Figure \ref{fig_framework}(a), the prompt set is shared across all images.
Consequently, the text noise generator can only take text features as input, ignoring the image $x$.
MLP can serve as this function $f_t(\mathcal{P})$. 

In another implementation, we set $\mu_t,\sigma_t$ as learnable embeddings.
Since $\mu_t$ and $\sigma_t$ have no functional relationship with $x$ and $\mathcal{P}$, $f_t(x, \mathcal{P})$ is effectively a constant function.
% introduce f_theta to mu and sigma 
% second phase sample
% image cross text parameter embedding
% figure framework

\section{Experiments}
\begin{figure*}
    \centering
    \includegraphics[width=\linewidth]{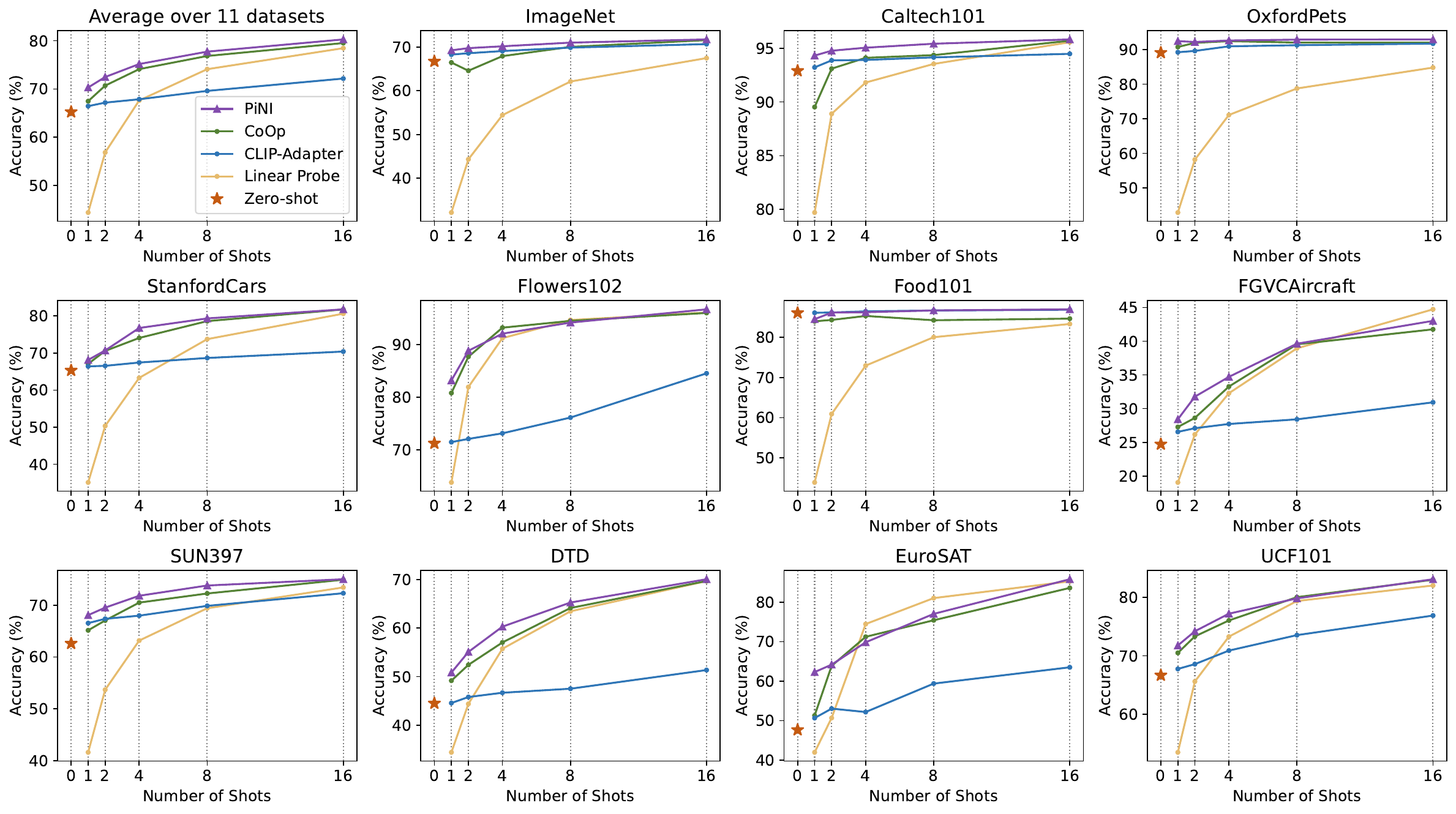}
    \caption{Performance of few-shot learning across 11 datasets. 
    In the top-left subplot, the results are averaged over 11 datasets.
    PiNI shows better performance compared to baselines, especially under conditions with fewer shots.
    }
    \label{fig_fewshot}
\end{figure*}
\subsection{Benchmark Settings}
\paragraph{\bf Datasets}
To evaluate the performance of PiNI, 11 datasets covering a wide range of visual concepts are selected. 
They include two generic object datasets, ImageNet \cite{ImageNet} and Caltech101 \cite{Caltech101};
five fine-grained datasets, OxfordPets \cite{OxfordPets}, StanfordCars \cite{StanfordCars}, Flowers102 \cite{Flowers102}, Food101 \cite{Food101} and FGVCAircraft \cite{FGVCAircraft}, which contain fine-grained categories of pets, cars, flowers, food and aircraft, respectively.
The other datasets are scene recognition dataset SUN397 \cite{SUN397}, action recognition dataset UCF101 \cite{UCF101}, describable textures dataset DTD \cite{DTD} and EuroSAT \cite{EuroSAT} which contains satellite images.
These datasets are abbreviated as Net, Caltech, Pets, Cars, Flowers, Food, Air, SUN, UCF, DTD and SAT. 

\begin{table}[t]
    \centering
    \renewcommand\arraystretch{1.0}
    \begin{adjustbox}{width=0.48\textwidth,center}
    \begin{tabular}{c|c|cccc}
        \toprule
        %\rowcolor{gray!40}
        \textbf{Loc.} & \textbf{Net.} &  \textbf{ImageNet} &  \textbf{Caltech} &  \textbf{Pets} & \textbf{Food} \\
        \hline
        & MLP & 67.43 &  94.52&  90.92& 85.39 \\
         \textbf{Image} & CNN &  67.54&  93.67&  89.51& 86.03 \\ 
         & CA &  67.19&  94.36&  90.48& 85.12 \\
         \hline
         \textbf{Visual} & MLP &  69.51&  95.41&  \textbf{92.25}& 85.91 \\
         \rowcolor{gray!20}
         \cellcolor{white}  \textbf{Feature} & CA &  \textbf{70.93}&  \textbf{95.82}&  91.99& \textbf{86.12}\\
         \hline
         \hline
         \rowcolor{gray!20}
        \cellcolor{white} \textbf{PE} & LE & \textbf{71.64}& \textbf{95.61}& \textbf{92.69}&86.98 \\
        \hline
        \textbf{Text} & LE & 70.74& 95.45& 92.25&\textbf{87.01}\\
        \textbf{Feature} & MLP & 71.14& 95.13& 91.44&85.29 \\
        \bottomrule
    \end{tabular}
    \end{adjustbox}
    \caption{Performance of PiNI adopting different injection locations and network architectures. 
    The number of shots is 16.
    PE is short for prompt embedding;
    CA is short for Cross-attention; 
    LE is short for learnable embedding.
    Visual Feature + CA and PE + LE show the best performance at the visual and text ends, respectively.
    }
    \label{tab_explore}
\end{table}
\begin{table}[t]
  \begin{adjustbox}{width=0.47\textwidth,center}
  \centering
  \begin{tabular}{c|cccc}
    \toprule
    \multirow{2}{*}{Shot}&Linear& CoOp& CLIP-Adapter&PiNI\\
    & Probe & (IJCV'22) & (IJCV'24) & (Ours)\\
    \midrule
    0& 65.23& 65.23& 65.23&65.23\\
    1& 44.36& 67.46& 66.44&\textbf{70.30}\\
    2& 56.82& 70.69& 67.16&\textbf{72.47}\\
    4& 67.61& 74.11& 67.86&\textbf{75.17}\\
 8& 74.10& 76.84& 69.59&\textbf{77.74}\\
 16& 78.46& 79.53& 72.16&\textbf{80.28}\\
  \bottomrule
  \end{tabular}
  \end{adjustbox}
  \caption{The performance averaged across 11 datasets under different shot settings. }
  \label{tab_dataset_avg}
\end{table}

\paragraph{\bf Baseline Methods} 
PiNI is compared with four baseline methods: Zero-shot CLIP (ZS) \cite{CLIP}, Linear Probe (LP), CoOp \cite{CoOp}, and CLIP-Adapter (CAda) \cite{CLIP-Adapter}. 
For Linear Probe, logistic regression is applied to classify the visual features of images. 
CoOp substitutes the word embeddings of prompt template with learnable parameters. 
CLIP-Adapter adds an extra module at the end of CLIP. The default parameter configurations are used for these baselines. 

\paragraph{\bf Training Details}
In the few-shot learning experiments, the train dataset is randomly sampled with 1, 2, 4, 8, and 16 shots per category.
The model is tested on all data in the test dataset. 
To ensure the fairness of the experiment, the best-performing ViT-B/16 is selected as the visual encoder unless otherwise noted.
The noise sample number $m$ in Eq. (\ref{eqn_loss}) is set to 1.
More details can be found in Appendix A.

\subsection{Exploring the Framework of PiNI} \label{subsec_explore}
Based on the discussion in Section \ref{subsec_noise_fine_tune}, we explore different noise incorporation approaches and network architectures for noise distribution parameter estimation.
We employ a 10-layer convolutional network with residual connections as the CNN architecture.
For the cross-attention architecture, we input the text feature as the key and value. The query is the image's spatial feature when injecting into raw images, as shown in Figure \ref{fig_framework}(e). 
When visual features are injected, the query is the feature itself.

Table \ref{tab_explore} presents the performance of different combinations.
For visual noise, injecting into visual feature with the cross-attention architecture yields the best performance.
Regarding text noise, it is more effective to inject into the prompt embeddings using learnable embeddings as the generator.
In the following experiments, we use these two combinations as visual and text noise generators as the default.

\paragraph{\bf Visualization}
To understand how noise works, we visualize the noise injected into raw images, as shown in Figure \ref{fig_visual}.
On the one hand, the injected noise can alleviate dataset bias, making objects in images resemble more common categories, such as the barrel in the first column and the gramophone in the third column.
On the other hand, the noise can interfere with regions unrelated to the category, thereby simplifying the classification task. For instance, the vegetation and ball around the dog in the last column.
Since the given prompt contains only a single category name, masking irrelevant regions in the image can enhance the alignment between the image and prompt embeddings. 

\begin{table}[t]
    \centering
    \begin{tabularx}{0.47\textwidth}{l
    >{\centering\arraybackslash}p{1.2cm}
    >{\centering\arraybackslash}X
    >{\centering\arraybackslash}X
    >{\centering\arraybackslash}X
    >{\centering\arraybackslash}X
    >{\centering\arraybackslash}X}
        \toprule
        \multirow{2}{*}{\textbf{Method}} & \textbf{Source} & \multicolumn{4}{c}{\textbf{Target}} & \multirow{2}{*}{\textbf{Avg.}} \\
         \cmidrule{3-6}
        & ImageNet & -V2 & -S & -A & -R & \\ 
        \midrule
         ZS&  66.72 &  60.81&  46.12&  47.70&  74.01&  59.07 \\
         CoOp & 71.58 &   63.76&  46.34&  48.21&  73.18&  60.61\\
         CAda&  70.67 &   63.35&  47.42&  \textbf{48.82} &  73.94&  60.84 \\
         PiNI& \textbf{71.74} &   \textbf{64.40}&  \textbf{48.25}&   48.57&  \textbf{74.39} &  \textbf{61.47} \\
         \bottomrule
    \end{tabularx}
    \caption{Comparison of PiNI with other methods on robustness to distribution shift.}
    \label{tab_domain_gen}
\end{table}
\begin{table}[t]
    %\begin{adjustbox}{width=0.47\textwidth,center}
    \begin{tabularx}{0.47\textwidth}{p{1.2cm} X X X X X}
         \toprule
         \multirow{2}{*}{\textbf{Dataset}} &  \multirow{2}{*}{\textbf{Method}} &  \multirow{2}{*}{\textbf{RN50}} &  \multirow{2}{*}{\textbf{RN101}} & \textbf{ViT-} & \textbf{ViT-} \\
          &   &   & & \textbf{B/32} & \textbf{B/16} \\
         \midrule
         \multirow{4}{*}{\textbf{ImageNet}} &  ZS&  58.19&  61.26&  62.04& 66.72 \\
         &  CoOp&  63.06&  71.68&  66.62& 71.58 \\
         &  CAda&  63.20 &  70.65&  65.81 & 70.66 \\
         &  PiNI&  \textbf{64.13}&  \textbf{72.01} &  \textbf{66.84} & \textbf{71.74} \\
         \midrule
         \multirow{4}{*}{\textbf{Caltech}} &  ZS&  85.92&  89.65&  91.11& 92.90 \\
         &  CoOp&  92.37&  95.70 &  95.13& 95.70 \\
         &  CAda&  90.50 &  94.48&  93.55& 94.48 \\
         &  PiNI&  \textbf{93.02} &  \textbf{95.86} &  \textbf{95.42} & \textbf{95.82} \\
         \bottomrule
    \end{tabularx}
    %\end{adjustbox}
    \caption{Performance of PiNI in different visual backbones.}
    \label{tab_visual_backbone}
\end{table}
\begin{figure}[t]
    \centering
    \includegraphics[width=\linewidth]{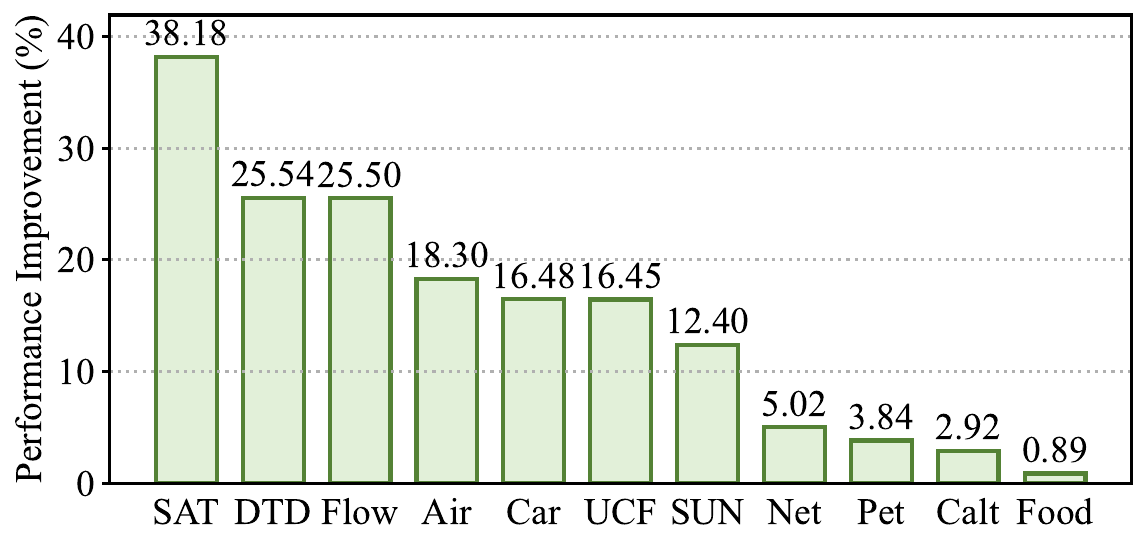}
    \caption{Performance improvement of PiNI compared to Zero-shot CLIP.
    The number of shots is 16. 
    }
    \label{fig_improvement}
\end{figure}

\subsection{Few-Shot Learning}
The performance of different methods across 11 datasets is shown in Figure \ref{fig_fewshot}. 
From the figure, it is clear that PiNI has significant performance improvement.  
The results averaged across 11 datasets are provided in Table \ref{tab_dataset_avg}.
It can be observed that the performance improvement is more pronounced with fewer shots.

The bar chart shown in Figure \ref{fig_improvement} illustrates the performance improvement of PiNI compared to Zero-shot CLIP.
The maximum performance improvement is 38.18\% on EuroSAT. 
On ImageNet, Caltech101, OxfordPets, and Food101, the performance improvements are relatively small as these categories are extensively contained in the corpus during pre-training stage.

Compared to Linear Probe, PiNI performs worse on FGVCAircraft.
The aircraft names in FGVCAircraft are rarely included in the corpus of the pre-training phase, such as "Boeing 737".
These specialized terms make prompt embeddings inaccurate in representing the fine-grained categories of aircraft.
However, Linear Probe only utilizes the visual encoder for classification, which eliminates the interference of inaccurate prompt embeddings.

\subsection{Domain Generalization}
In domain generalization experiments, the model is trained on the source dataset ImageNet, and tested on target datasets. 
The target datasets are four variants of ImageNet, namely, ImageNetV2 \cite{ImageNetV2}, ImageNet-Sketch \cite{ImageNet-Sketch}, ImageNet-A \cite{ImageNet-A} and ImageNet-R \cite{ImageNet-R}.
Detailed results are shown in Table \ref{tab_domain_gen}. 
PiNI outperforms other methods except on ImageNet-A, exhibiting strong robustness against distribution shift. 
  
\subsection{Ablation on Visual Backbones}
To investigate the impact of visual backbones in CLIP, we conduct 16-shot experiments, as shown in Table \ref{tab_visual_backbone}.
These visual backbones include RN-50, RN-101, ViT-B/32, and ViT-B/16. 
On the two generic object datasets, ImageNet and Caltech, PiNI consistently demonstrates superior performance across different visual backbones.

\section{Conclusion}
In this work, we propose PiNI, a noise-based fine-tuning method towards vision-language alignment. 
It implies a new scheme to learn beneficial noise distribution.
In our experiments, we demonstrate that PiNI outperforms existing methods.
 Our work can be further extended by refining the task entropy in Eq. (\ref{eqn_HT}).
In this work, we select prompt-based image classification as the primary task. 
However, CLIP has a wide range of applications,  such as Visual Question Answering (VQA), object detection, and image generation. 
Through defining task entropy on these tasks, the beneficial noise can be generated to simplify their complexity.

\bibliography{aaai25}

\clearpage

\appendix

\section{Additional Implementation Details}  \label{appendix_details}
\subsection{Datasets}
We use 11 datasets for few-shot learning experiments and four datasets for domain generation experiments.
In the domain generation experiments, we evaluate the model only on the four datasets with distribution shift, thus ignoring the training sets.
Generally, we follow the same dataset processing procedures as those in CoOp \cite{CoOp}.
Details of these datasets are provided in Table \ref{tab_dataset}.

\subsection{Prompts}
We use the default prompts from \cite{CLIP} for our experiments. 
For datasets of generic objects or scenes, the typical form ``A photo of a [class].'' is adopted. For other datasets in specific domains, additional context is added to the prompts. For instance, the prompt in OxfordFlowers is ``A photo of a [class], a type of flower.''
The corresponding  prompts are shown in Table \ref{tab_dataset}.

The prompt that we select for PiNI differs slightly from hand-crafted prompts in CLIP \cite{CLIP} and CLIP-Adapter \cite{CLIP-Adapter}.
The difference lies in placing all the category names at the end of the template to avoid splitting the prompt template into two parts.
This approach simplifies adding noise into the hand-crafted prompts.
Additionally, we insert random words at the beginning of the prompt to ensure the same number of learnable word embeddings as CoOp.

\subsection{Training}
In all experiments, a batch size of 32 is used. 
For ImageNet containing 1000 categories, the number of epochs is set to 100, while for the other datasets, it is set to 200. 
The SGD optimizer with a Cosine Annealing Learning Rate is employed.
For ImageNet, the initial learning rate is 0.001. For other datasets, the learning rate is between 0.001 to 0.003.
We also apply a constant warm-up phase with a learning rate of 1e-5, which lasts for 1 epoch.
All experiments are conducted using a single NVIDIA 4090 GPU.

\section{Additional Results}  \label{appendix_results}
\subsection{Few-Shot Learning}
We provide detailed few-shot learning results for 11 datasets in Table \ref{tab_16_shots}.
In this table, PiNI is compared with Zero-shot CLIP, Linear Probe, CoOp, and CLIP-Adapter in the 16-shot learning.
In Table \ref{vit-l}, we apply ViT-L-14@336px to PiNI and other baselines. PiNI also enhances the performance of strong CLIP models, particularly when there are very few training samples. 

\subsection{Visualization of Noise}
\begin{figure*}
    \centering
    \includegraphics[width=\linewidth]{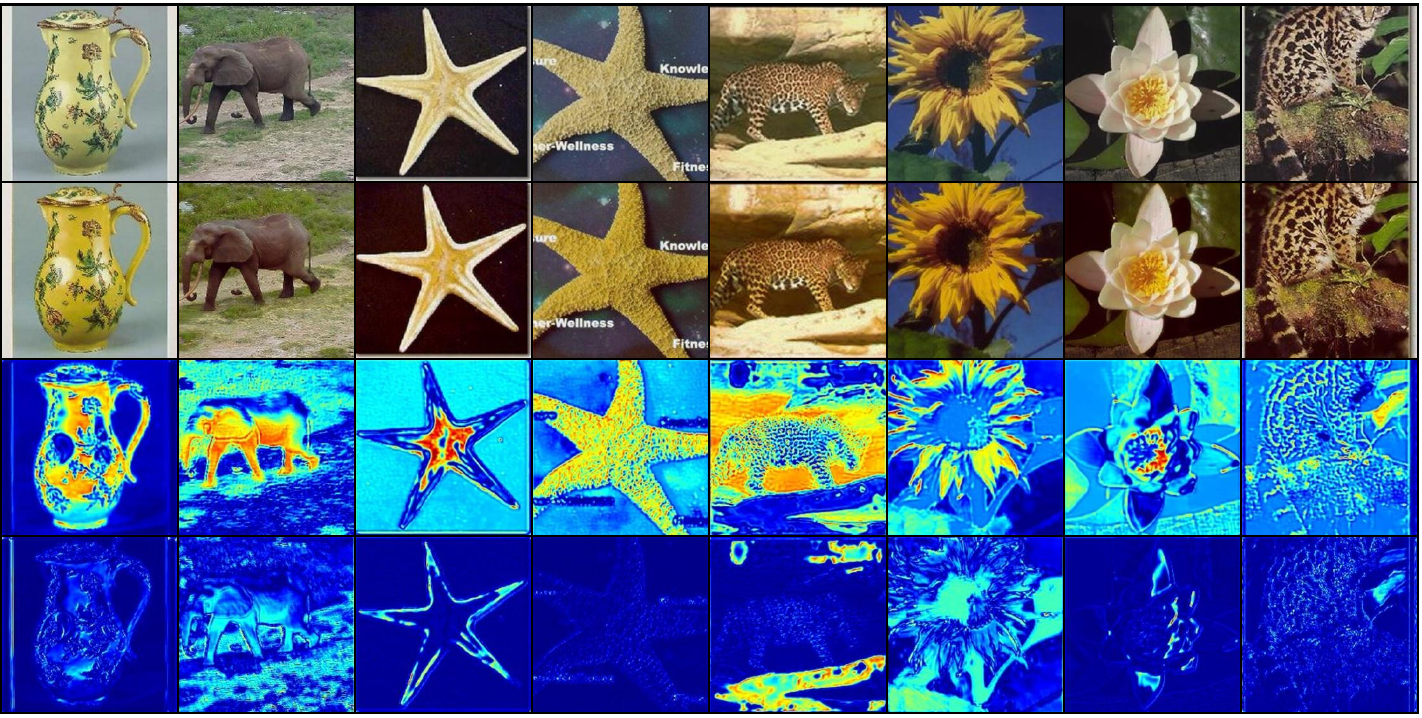}
    \caption{Visualization of the noise injected into raw images from Caltech101. 
    }
    \label{fig_visual_caltech}
\end{figure*}

\begin{figure*}
    \centering
    \includegraphics[width=\linewidth]{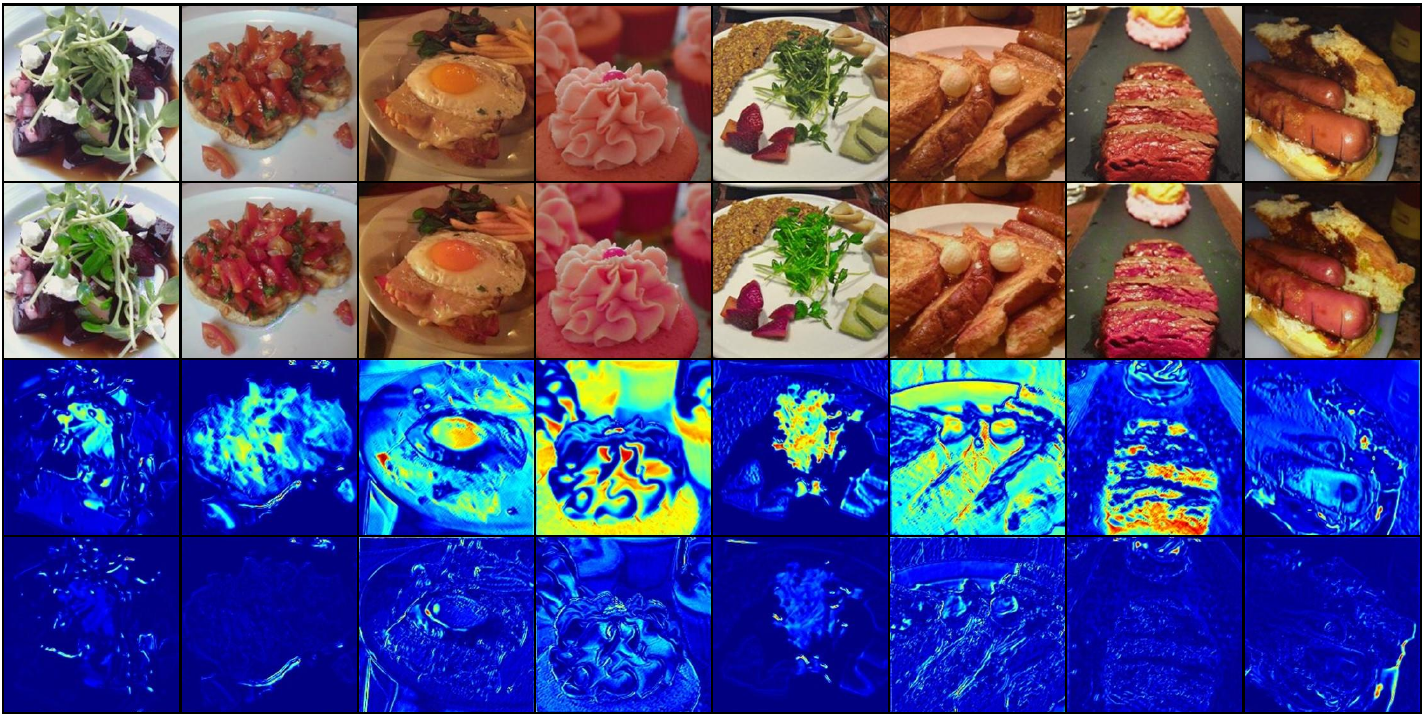}
    \caption{Visualization of the noise injected into raw images from Food101. 
    }
    \label{fig_visual_food}
\end{figure*}

\begin{figure*}
    \centering
    \includegraphics[width=\linewidth]{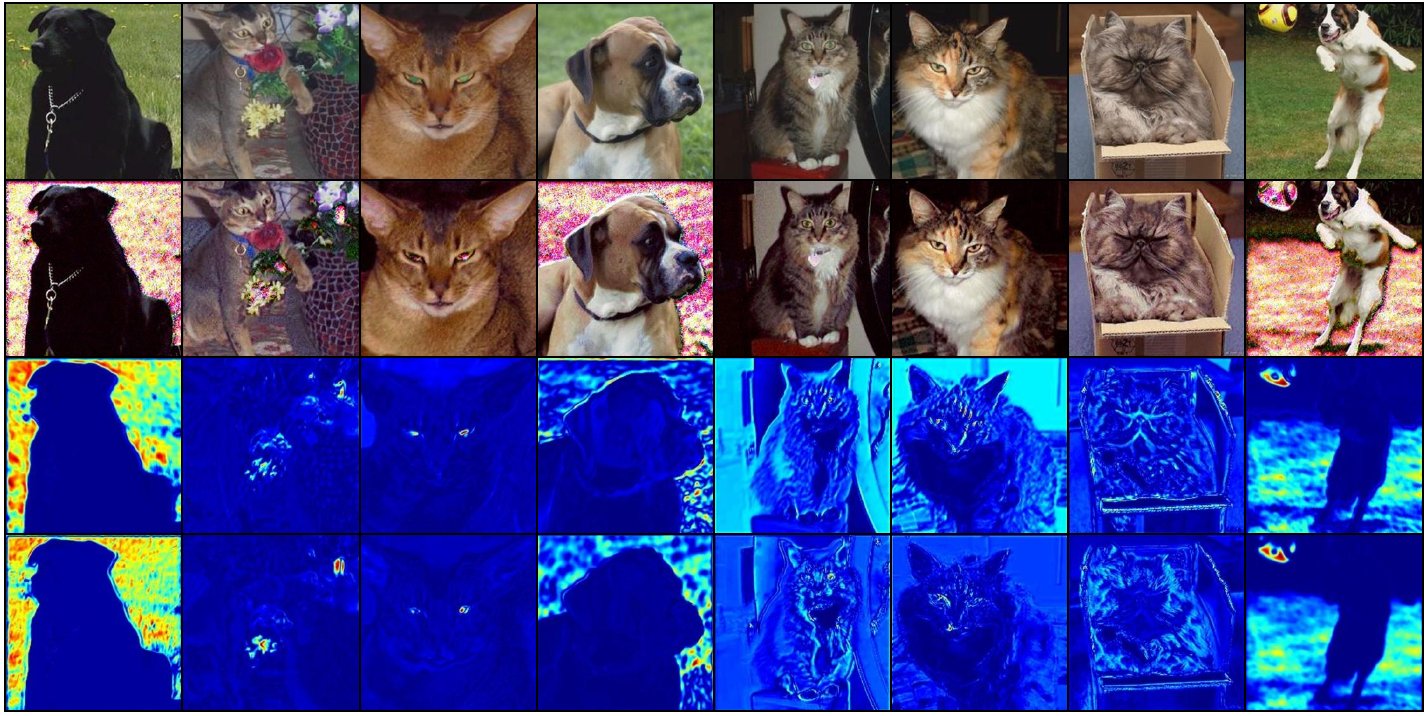}
    \caption{Visualization of the noise injected into raw images from OxfordPets. 
    }
    \label{fig_visual_pets}
\end{figure*}

We visualize the noise injected into raw images from Caltech101, Food101, and OxfordPets in Figures \ref{fig_visual_caltech}, \ref{fig_visual_food}, and \ref{fig_visual_pets}, respectively.
The first row shows the raw images. The second row displays the noise-injected images. The third and fourth rows present heatmaps of the mean $\mu$ and variance $\sigma$ for each pixel, respectively.

\subsection{Interpretation of the Learned Prompt Distribution}
In our proposed method PiNI, random noise is injected into the word embeddings of the prompt template, so the template itself also follows a probability distribution.
However, interpreting the learned distribution of word embeddings is challenging, because they exist in continuous space, whereas the embeddings of real words are in discrete space.
We address this issue by finding the nearest embeddings of real words in the vocabulary.
Specifically, we sample the noise from the distribution 100 times and add it to the prompt template. 
For each sample, we record five words nearest to the word embedding at each position of the prompt template.
Table \ref{tab_word} shows the most frequently occurring words at each position of the prompt template.
Note that the vocabulary includes many subwords, e.g., ``pic" is a subword of ``picture". 
By adding noise to the prompt, we find that the breadth of semantics expressed by the words in the prompt template has increased. 
For example, the words ``doing", ``do", and ``does" express the same meaning in different grammatical contexts.
Additionally, ``photo", ``photos", ``pic" and ``picture" also have similar meanings.
In conclusion, these words with similar meanings demonstrate that the obtained prompt distribution has richer semantics, confirming the effectiveness of our proposed PiNI.

\begin{table}
    \centering
    \begin{tabular}{ccccc}
    \toprule
         Dataset&  Method&  shot=2&  shot=8& shot=16 \\
    \midrule
         \multirow{4}{*}{Caltech} &  LP&  91.52&  95.05& 96.31 \\
         &  CoOp&  95.82&  96.43& 97.12 \\
         &  CAda&  94.68&  96.10& 96.63 \\
         &  PiNI&  \textbf{96.27}&  \textbf{96.87}& \textbf{97.28}\\
    \midrule
         \multirow{4}{*}{DTD} &  LP&  42.85&  65.13& 72.1 \\
         &  CoOp&  58.03&  \textbf{70.74}& \textbf{73.75}\\
         &  CAda&  56.5&  59.87& 64.36 \\
         &  PiNI&  \textbf{59.39}&  70.56& 73.58 \\
    \midrule
         \multirow{4}{*}{OxfordPets} &  LP&  66.53&  84.08& 89.56 \\
         & CoOp& 93.67& 94.19&94.35 \\
         & CAda& 94.03& 94.33&94.52 \\
         & PiNI& \textbf{94.41}& \textbf{95.01}&\textbf{95.12}\\
    \midrule
         \multirow{4}{*}{SUN397} & LP& 56.02& 71.34&74.92 \\
         & CoOp& 69.44& 76.35&\textbf{78.40}\\
         & CAda& 71.79& 74.85&77.01 \\
         & PiNI& \textbf{73.22}& \textbf{76.74}&78.16 \\
    \bottomrule
    \end{tabular}
    \caption{Comparison of PiNI with other methods using the visual backbone of ViT-L-14@336px.}
    \label{vit-l}
\end{table}

\begin{table*}
    \centering
    \begin{tabular}{lcccc}
         \toprule
         Dataset&  Classes&  Train Size&  Test Size& Prompt \\
         \midrule
         ImageNet&  1,000&  1,281,167&  50,000& ``a photo of a [class]''\\
         Caltech101&  100&  4,128&  2,465& ``a photo of a [class]''\\
         OxfordPets&  37&  2,944&  3,669& ``a type of pet, a photo of a [class]''\\
         StanfordCars&  196&  6,509&  8,041& ``a photo of a [class]''\\
         Flowers102&  102&  4,093&  2,463 & ``a type of flower, a photo of a [class]''\\
         Food101&  101&  50,500&  30,300& ``a type of food, a photo of [class]''\\
         FGVCAircraft&  100&  3,334&  3,333& ``a type of aircraft, a photo of a [class]''\\
         SUN397&  397&  15,880&  19,850& ``a photo of a [class]''\\
         DTD&  47&  2,820&  1,692& ``texture [class]''\\
 EuroSAT& 10& 13,500& 8,100&``a centered satellite photo of [class]''\\
 UCF101& 101& 7,639& 3,783&``a photo of a person doing [class]''\\
 \midrule
 ImageNetV2& 1,000& \textendash& 10,000&``a photo of a [class]''\\
 ImageNet-Sketch & 1,000& \textendash& 50,889&``a photo of a [class]''\\
 ImageNet-A& 200& \textendash& 7,500&``a photo of a [class]''\\
 ImageNet-R& 200& \textendash& 30,000&``a photo of a [class]''\\
 \bottomrule
    \end{tabular}
    \caption{The details of the datasets and their corresponding  prompts.}
    \label{tab_dataset}
\end{table*}

\begin{table*}
  \centering
   \begin{tabular}{lccccccccccc}
    \toprule
         Method& ImageNet&  Caltech& Pets& Cars& Flowers& Food& Aircraft& SUN& DTD& SAT& UCF \\
    \midrule
         Zero-shot &  66.72&  92.90&  89.13&  65.31&  71.21&  86.11&  24.72&  62.60&  44.50& 47.65&66.66
\\
         Linear Probe&  67.47&  95.56&  84.82&  80.62&  96.10&  83.35&  \textbf{44.73}&  73.42&  69.66& 85.30&82.04
\\
         CoOp&  71.58&  95.70&  92.04&  81.76&  96.02&  84.68&  41.76&  74.91&  69.68& 83.63&83.02
\\
         CLIP-Adapter&  70.67&  94.48&  91.76&  70.40&  84.53&  86.90&  30.93&  72.30&  51.35& 63.51&76.89
\\
         PiNI&  \textbf{71.74}&  \textbf{95.82}&  \textbf{92.97}&  \textbf{81.79}&  \textbf{96.71}&  \textbf{87.00}&  43.02&  \textbf{75.00}&  \textbf{70.04}& \textbf{85.83}&\textbf{83.11}
\\
     \bottomrule
    \end{tabular}
    \caption{Comparison of PiNI with other methods in 16-shot learning.
  The names of these datasets are abbreviated.}
  \label{tab_16_shots}
\end{table*}

\begin{table*}
    \centering
    \begin{tabular}{lcccccc}
         \toprule
         Dataset&  Position&  1&  2&  3&  4& 5\\
         \midrule
         \multirow{9}{*}{OxfordPets}&  1&  a(100)&  an (99)&  the (87)&  my (54)& sundaywithmarsha (46)
\\
         &  2&  type(100)&  0 (83)&  types (79)&  1 (51)& pational (30)
\\
         &  3&  of(100)&  in (65)&  to (63)&  on (52)& 0 (51)
\\
         &  4&  pet (100)&  0 (99)&  pets (75)&  1 (45)& 2 (39)
\\
         &  5&  , (100)&  ., (60)&  ), (57)&  . (30)& ! (30)
\\
         &  6&  a (100)&  an (94)&  flyeagles (50)&  the (50)& 0 (34)
\\
         &  7&  photo (100)&  0 (84)&  1 (65)&  2 (45)& coscino (44)
\\
         &  8&  of (100)&  to (100)&  by (49)&  in (49)& with (40)
\\
         &  9&  a (100)&  an (79)&  sundaywithmarsha (51)&  0 (47)& 1 (28)
\\
        \midrule
         \multirow{6}{*}{UCF101}& 1& a (100)& an (84)& 0 (81)& the (56)&1 (55)
        \\
         & 2& photo (100)& photos (96)& pic (75)& 0 (74)&picture (39)
        \\
         & 3& of (100)& to (86)& for (55)& , (52)&at (28)
        \\
         & 4& a (100)& an (58)& the (58)& instaweather (53)&0 (49)
        \\
         & 5& person (100)& 0 (75)& people (57)& 1 (53)&pational (35)
        \\
         & 6& doing (100)& 0 (69)& do (63)& does (39)&1 (39)
        \\
        \bottomrule
    \end{tabular}
    \caption{The most frequently occurring words at each position in the prompt templates in OxfordPets and UCF101.
    The frequency of occurrence for each word is provided in parentheses.}
    \label{tab_word}
\end{table*}

\end{document}